\documentclass[sigconf]{acmart}

\usepackage{booktabs} 
\usepackage{xspace} 
\usepackage{bbm}

\newtheorem{theorem}{Theorem}[section]
\newtheorem{proposition}[theorem]{Proposition}
\newtheorem{remark}{Remark}[section]

\newtheorem{definition}{Definition}[section]
\newtheorem{corollary}[theorem]{Corollary}

\usepackage{algorithm}
\usepackage{algorithmic}

\newcommand{\E}{\mathop{\mathbb E}}
\newcommand{\Tr}{\operatorname{Tr}}

\newcommand{\N}{\mathcal{N}}

\newcommand{\vect}{\mathrm{vec}}

\def\x{{\mathbf x}}

\DeclareMathOperator*{\argmin}{arg \, min}

\renewcommand{\t}[1]{\mathrm{T}#1}
\renewcommand{\d}[1]{\;\mathrm{d}#1}
\renewcommand{\t}[1]{\mathrm{T}#1}

\setcopyright{rightsretained}

\acmConference[]{A.I Square Working Paper}{March 2019}{France}
\acmYear{2019}
\copyrightyear{2019}

\begin{document}
\title{BCMA-ES II: revisiting Bayesian CMA-ES}


\author{Eric Benhamou}
\affiliation{%
  \institution{A.I Square Connect and Lamsade, France}
}
\email{eric.benhamou@aisquareconnect.com}

\author{David Saltiel}
\affiliation{%
  \institution{A.I Square Connect and LISIC, France}
}
\email{david.saltiel@aisquareconnect.com}

\author{Beatrice Guez}
\affiliation{
  \institution{A.I Square Connect, France}
}
\email{beatrice.guez@aisquareconnect.com}

\author{Nicolas Paris}
\affiliation{
  \institution{A.I Square Connect, France}
}
\email{nicolas.paris@aisquareconnect.com}

\renewcommand{\shortauthors}{E. Benhamou, D. Saltiel, B. Guez, N. Paris}

\begin{abstract}
This paper revisits the Bayesian CMA-ES and provides updates for normal Wishart. It emphasizes the difference between a normal and normal inverse Wishart prior. After some computation, we prove that the only difference relies surprisingly in the expected covariance. We prove that the expected covariance should be lower in the normal Wishart prior model because of the convexity of the inverse. We present a mixture model that generalizes both normal Wishart and normal inverse Wishart model. We finally present various numerical experiments to compare both methods as well as the generalized method.
\end{abstract}

%
%



\maketitle

\section{Introduction}
Bayesian statistics have revolutionized statistics like quantum mechanics have done for Newtonian mechanism. Like the latter, the usual frequentist statistics can be seen as a particular asymptotic case of the former. Indeed, the Cox Jaynes theorem (\cite{Cox_1946}) proves that under the four axiomatic assumptions given by:
\begin{itemize}
\item plausibility degrees are represented by real numbers (continuity of method), 
\item none of the possible data should be ignored  (no retention)
\item these values follow usual common sense rule as stated by the well known Laplace formula: 
\textit{the probability theory is truly the common sense represented in calculus} (common sense), 
\item and states of equivalent knowledge should have equivalent degree of plausibility (consistency),
\end{itemize}
then, there exists a probability measure defined up to a monotonous function such that it follows the usual probability calculus and the fundamental rule of Bayes, that is:
\begin{equation}\label{eq:symmetric_relationship}
\mathbb{P}(H,  D) = \mathbb{P}(H | D) \mathbb{P}(D) = \mathbb{P}(D| H) \mathbb{P}(H)
\end{equation}
where $H$ and $D$ are two members of the implied $\sigma-$algebra. 
The letters are not by chance. $H$ stands for the hypothesis, which can be interpreted as an hypothesis on the parameters, while $D$ stands for data. 

The usual frequentist probabilities states that the probability of an observation $\mathbb{P}(D)$ is given certain hypothesis $H$ on the state of the world. However, as the equation
\eqref{eq:symmetric_relationship} is completely symmetric, nothing hinders us to change our point of view and state the inverse question. Given an observation of a data $D$, what is the plausibility of the hypothesis $H$. The Bayes rules trivially answers this question: 
\begin{equation}\label{eq:Bayes_rules1}
\mathbb{P}(H| D) = \mathbb{P}(D|H) \frac{\mathbb{P}(H)} {\mathbb{P}(D)} = \mathbb{P}(D|H) \mathbb{P}(H)
\end{equation}
or equivalently,
\begin{equation}\label{eq:Bayes_rules2}
\mathbb{P}(H|D) \propto \mathbb{P}(D|H) \mathbb{P}(H)
\end{equation}

In the above equation, $\mathbb{P}(H)$ is called the prior probability or simply the prior while the conditional probability $\mathbb{P}(H|D)$ is called the posterior probability or simply the posterior. 
There are a few remarks to be made. 
First of all, the prior is not necessarily independent of the knowledge of the experience, on the contrary, a prior is often determined with some knowledge of previous experience in order to make a meaningful choice. Second, prior and posterior are not necessarily related to a chronological order but rather to a logical order. 

After observing some data $D$, we revise the plausibility of $H$. it is interesting to see that the conditional probability $\mathbb{P}(D|H)$ considered as a function of $H$ is indeed a likelihood for $H$. The Cox Jaynes theorem as presented in \cite{jaynes_2003} gives the foundation for Bayesian calculus. Another important result is the De Finetti's theorem. 
Let us recall the definition of Infinite exchangeability.

\begin{definition} (Infinite exchangeability).
We say that $(x_1, x_2, . . .)$ is an infinitely exchangeable sequence of
random variables if, for any n, the joint probability $p(x_1, x_2, . . . , x_n)$ is invariant to permutation of the
indices. That is, for any permutation $\pi$,
$$
p(x_1, x_2, . . . , x_n) = p(x_{\pi1}, x_{\pi2}, . . . , x_{\pi n})
$$
\end{definition}

Equipped with this definition, the De Finetti's theorem as provided below states that exchangeable observations are conditionally independent relative to some latent variable.

\begin{theorem}
\label{theorem_finetti}
(De Finetti, 1930s). A sequence of random variables $(x_1, x_2, . . .)$ is infinitely exchangeable iff,
for all n,
$$
p(x_1, x_2, . . . , x_n) = \int \prod_{i=1}^n p(x_i|\theta) P(d\theta),
$$
for some measure P on $\theta$.
\end{theorem}

This representation theorem \ref{theorem_finetti} justifies the use of priors on parameters since for exchangeable data, there must exist a parameter $\theta$, a likelihood $p(x|\theta)$ and a distribution $\pi$ on $\theta$. A proof of De Finetti theorem is for instance given in~\cite{Schervish_1996} (section 1.5). We will see that this Bayesian setting gives a powerful framework for revisiting black box optimization that is introduced below.

\section{Black box optimization}
We assume that we have a real value $p$-dimensional function $f : \mathbb{R}^p \rightarrow \mathbb{R}$. We examine the following optimization program:
\begin{equation}
\underset{x \in \mathbb{R}^p }{\min} f(x)
\end{equation}
In contrast to traditional convex optimization theory, we do not assume that $f$ is convex, neither continuous nor admits a global minimum. We are interested in the so called \textit{Black box optimization} (BBO) settings where we only have access to the function $f$ and nothing else. By nothing else, we mean we can not for instance compute gradient. A practical way to do optimization in this very general and minimal setting is to do evolutionary optimization and in particular use the covariance matrix adaptation evolution strategy (CMA-ES) methodology. The  CMA-ES~\cite{HansenOstermeier_2001} is arguably one of the most powerful real-valued derivative-free optimization algorithms, finding many applications in machine learning. It is a state-of-the-art optimizer for continuous black-box functions as shown by the various benchmarks of the COCO (\href{http://coco.gforge.inria.fr/}{COmparing Continuous Optimisers}) INRIA platform for ill-posed functions. It has led to a large number of papers and articles and we refer the interested reader to~\cite{HansenOstermeier_2001,Auger_2004,Igel_2007,Auger_2009,Hansen_2011,Auger_2012,Hansen_2014,Auger_2015,Auger_2016,Ollivier_2017} and~\cite{Hansen_2018} to cite a few.  

It has has been successfully applied in many unbiased performance comparisons and numerous real-world applications. In particular, in machine learning, it has been used for direct policy search in reinforcement learning and hyper-parameter tuning in supervised learning (~\cite{Gomez_2008},~\cite{Igel_2009a,Igel_2009b,Igel_2010}), and references therein, as well as hyperparameter optimization of deep neural networks ~\cite{Loshchilov_2016}.

In a nutshell, the ($\mu$ / $\lambda$) CMA-ES is an iterative black box optimization algorithm, that, in each of its iterations, samples $\lambda$ candidate solutions from a multivariate normal distribution, evaluates these solutions (sequentially or in parallel) retains $\mu$ candidates and adjusts the sampling distribution used for the next iteration to give higher probability to good samples. Each iteration can be individually seen as taking an initial guess or \emph{prior} for the multi variate parameters, namely the mean and the covariance, and after making an experiment by evaluating these sample points with the fit function updating the initial parameters accordingly. Although rethinking the CMA-ES in terms of a prior and posterior seems natural when coming over from Bayesian statistics, it is only recently that it has been explored ~\cite{BenhamouSaltiel_2019}. 

Historically, the CMA-ES has been developed heuristically. It was done mainly by conducting experimental research and validating intuitions empirically.

Research was done without much focus on theoretical foundations because of the apparent complexity of this algorithm. 
It was only recently that ~\cite{Akimoto_2010},~\cite{Glasmachers_2010} and~\cite{Ollivier_2017} made a breakthrough and provided a theoretical justification of CMA-ES updates thanks to information geometry. 
They proved that CMA-ES was performing a natural gradient descent in the Fisher information metric. 
The Bayesian formulation of the CMA-ES came effectively much later and has only been done sofar with the normal inverse Wishart prior. 

In this paper, we revisit the Bayesian CMA-ES formulation and show that there exists indeed an infinity of conjugate prior given by the convex combination of a normal Wishart and normal inverse Wishart Gaussian prior. We first prove that normal Wishart and normal inverse Wishart Gaussian priors have the same update equations except for the mean of the covariance matrix. We provide a theoretical argument to show that the inverse of a matrix should be lower than in the normal inverse Wishart Gaussian prior. We then introduce a new prior given by a mixture of normal Wishart and normal inverse Wishart Gaussian prior. Likewise, we derive the update equations. In section \ref{sec:experiments}, we finally give numerical results to compare all these methods.

\section{Conjugate priors}\label{sec:various_priors}
A key concept in Bayesian statistics is conjugate priors that makes the computation really easy and is described below.

\begin{definition}\label{sec:conjprior}
A prior distribution $\pi(\theta)$ is said to be a conjugate prior if the posterior distribution 
\begin{align}
\pi(\theta | x ) \propto   p(x | \theta) \pi(\theta) \label{eq:ExponentialFamilyPosterior}
\end{align}
remains in the same distribution family as the prior.
\end{definition}

At this stage, it is relevant to introduce exponential family distributions as this higher level of abstraction that encompasses the multi variate normal trivially solves the issue of founding conjugate priors. This will be very helpful for inferring conjugate priors for the multi variate Gaussian used in CMA-ES.

\begin{definition}
A distribution is said to belong to the exponential family if it can be written (in its canonical form) as:
\begin{equation}
p(\x | \eta)=h(\x)\exp(\eta \cdot T(\x)-A(\eta )),
\label{eq:ExponentialFamilyNaturalForm}
\end{equation}
where  $\eta$ is the natural parameter, $T(\x)$ is the sufficient statistic, $A(\eta)$ is log-partition function and  $h(\x)$ is the base measure. $\eta$ and  $T(\x)$ may be vector-valued. Here $a \cdot b$ denotes the inner product of $a$ and $b$.

\noindent The log-partition function is defined by the integral:
\begin{align}
A(\eta)\triangleq \log \int_{\mathcal{X}}{h(\x)\exp({\eta \cdot T(\x)}) \d x}.
\end{align}
Also, $\eta \in \Omega=\{\eta \in \mathbb{R}^m |A(\theta)< +\infty\}$ where $\Omega$ is the natural parameter space. Moreover, $\Omega$ is a convex set and $A(\cdot)$ is a convex function on $\Omega$.
\end{definition}

\begin{remark} \label{ex:exponentialfamily} 
Not surprisingly, the normal distribution $\N(\x;\mu,\Sigma)$ with mean $\mu\in\mathbb{R}^d$ and covariance matrix $\Sigma$ belongs to the exponential family but with a different parametrisation. Its exponential family form is given by:
\begin{subequations}
\label{eq:normalExpForm}
\begin{align}
\eta(\mu,\Sigma)&=\begin{bmatrix} \Sigma^{-1}\mu \\  \mathrm{vec}(\Sigma^{-1}) \end{bmatrix}, \qquad T(\x) =\begin{bmatrix} \x\\ \mathrm{vec}(-\frac{1}{2} \x\x^\t) \end{bmatrix} \label{eq:Natparams},\\
h(\x)&=(2\pi)^{-\frac{d}{2}}, \qquad  A(\eta(\mu,\Sigma)) =\frac{1}{2}\mu^\t \Sigma^{-1}\mu+\frac{1}{2}\log|\Sigma|. \label{eq:Natparams2}
\end{align} 
\end{subequations}
where in equations  \eqref{eq:Natparams}, the notation $\vect(\cdot)$ means we have vectorized the matrix, stacking each column on top of each other and hence can equivalently write for $a$ and $b$, two matrices, the trace result $\Tr(a^\t{}b)$ as the scalar product of their vectorization $\vect(a)\cdot \vect(b)$ (see \ref{proof:one_line}). We can remark the canonical parameters are very different from traditional (also called moment) parameters. We can notice that changing slightly the sufficient statistic $T(x)$ leads to change the corresponding canonical parameters $\eta$. In equation \eqref{eq:Natparams2}, the notation $|\Sigma|$ means the determinant of the matrix: $\det(\Sigma)$.
\end{remark}

\noindent For an exponential family distribution, it is particularly easy to form conjugate prior. 

\begin{proposition}\label{prop:conjugate_priors}
If the observations have a density of the exponential family form $p(x |\theta,\kappa) = h(x) \exp\Big(\eta(\theta,\kappa)^T T(x)-n A(\eta(\theta,\kappa))\Big)$, with $\kappa$ a set of hyper-parameters, then the prior with likelihood defined by $\pi(\theta) \propto \exp\left( \lambda_1\cdot \eta(\theta,\kappa) - \lambda_0 A(\eta(\theta,\kappa)) \right)$ with $\lambda\triangleq(\lambda_0, \lambda_1)$ is a conjugate prior.
\end{proposition}

The proof is given in appendix subsection \ref{proof:conjugate_priors}. As we can vary the parameterisation of the likelihood, we can obtain multiple conjugate priors. Because of the conjugacy, if the initial parameters of the multi variate Gaussian follows the prior, the posterior is the true distribution given the information $\mathcal{X}$ and stay in the same family making the update of the parameters really easy. Said differently, with conjugate prior, we make the optimal update.  

A consequence of proposition \ref{prop:conjugate_priors} is that the various conjugate priors of the multi variate normal that belong to the exponential family can be determined. This is the subject of the corollary below.

\begin{corollary}\label{cor:conjugate_priors_2}
The conjugate priors of the multi variate normal that belong to the exponential family are necessarily of the form :
\begin{itemize}
\item normal inverse Wishart distribution $NIW( \mu_0, \lambda_0,\nu_0, \Psi_0)$ if the multivariate normal is described in terms of its mean vector $\mu$ and covariance matrix $\Sigma$.
\item normal Wishart distribution $NW( \mu_0, \lambda_0, \nu_0, W_0)$ if the multivariate normal is described in terms of its mean vector $\mu$ and precision matrix $\Lambda$.
\end{itemize}
\end{corollary}
 
The proof is given in appendix subsection \ref{proof:conjugate_prior_normal}. As conjugate priors, the posterior of the two identified distributions of the corollary \ref{cor:conjugate_priors_2} are easy to derive and are given by the following proposition.

\begin{proposition}\label{prop:posterior_update}
For a likelihood of $n$ points $(x)_{i=1..n}$ distributed according to a multi variate normal distribution whose parameters are given by the priors below: 
\begin{enumerate}
\item the normal inverse Wishart distribution:\\ $NIW_0 = NIW( \mu_0, \lambda_0,\nu_0, \Psi_0)$
\item the normal Wishart distribution: $NW_0 = NW( \mu_0, \lambda_0, \nu_0, W_0)$ 
\item the mixture of a normal inverse and normal Wishart with same parameters: $w NIW_0 + (1-w) NW_0$ with $0\leq w \leq 1$
\end{enumerate}
The posterior is given by:
\begin{enumerate}
\item the normal inverse Wishart distribution 

\begin{equation}\label{eq:update_posterior_niw}
NIW_1 = NIW\left( \frac{\lambda_0 \, \mu_0 + n \overline{x}}{\lambda_0+n}, \lambda_0 + n, \nu_0 + n, \Psi_0 + n C +n D \right)
\end{equation}

\item the normal Wishart distribution 

\begin{equation}\label{eq:update_posterior_nw}
NW_1 = NW\left( \frac{\lambda_0 \, \mu_0 + n \overline{x}}{\lambda_0+n}, \lambda_0 + n, \nu_0 + n, \left( W_0 + n C + n D \right)^{-1} \right)
\end{equation}

\item the mixture of a normal inverse and normal Wishart with same parameters:  $w NIW_1 + (1-w) NW_1$
\end{enumerate}
where \,$\overline{x} = 1/n \sum_{i=1}^n x_i$ is the sample mean, \, $C =  1/n \sum_{i=1}^n (x_i - \overline{x})(x_i - \overline{x})^T$ the sample covariance and $D= \frac{\lambda_0 \, n}{n(\lambda_0+n)} ( \overline{x}-\mu_0) ( \overline{x}-\mu_0)^T$.
\end{proposition}

The proof is given in appendix subsection \ref{proof:conjugate_prior_computation}.

\section{Algorithm}\label{sec:algorithm}
The idea behind the algorithm is at each step to make use the previous iteration posterior as a prior, draw the likelihood and then update according to proposition \eqref{prop:posterior_update} the posterior. In full generality, the prior is a distribution, so we would need to do a Monte Carlo of Monte Carlo. But in order to reduce the variance by this Monte Carlo of Monte Carlo, we make the simplification to use the mean value of the prior distribution. These values are given as follows:

\begin{enumerate}
\item for the normal inverse Wishart distribution, $\hat{\mu}= \E[\mu] = \mu_{n}$ and $\hat{\Sigma}=\E[\Sigma] = \Psi_n/(v_n - p -1)$
\item for the normal Wishart distribution, $\hat{\mu}= \E[\mu] = \mu_{n}$ and $\hat{\Sigma}=\E[\Lambda^{-1}] = \Psi_n/v_n$ for $\Psi_n= W_n^{-1}$.
\item for the  $w$ mixture of the normal inverse and normal Wishart with same parameters, $\hat{\mu}= \E[\mu] = \mu_{n}$ and 
$\hat{\Sigma}=\E[\Sigma] =\frac{ v_n - p -1  +w p + w } {v_n(v_n - p -1)} \Psi_n$
\end{enumerate}

It is obvious that the expected value of the covariance matrix of the normal inverse Wishart $\hat{\Sigma}=\E[\Sigma]$ should be above the one of the normal Wishart distribution as the inverse of a matrix $Inv : S \rightarrow S^{-1}$ is a convex function in the domain $\mathcal{S}^p_{++}$ of symmetric definite positive matrices. A proof is given in \ref{proof:inverse_matrix}.
To recover the true minimum, we design two strategies. 
\begin{itemize}
\item we design a strategy where we rebuild our normal distribution but using sorted information of our $X$'s weighted by their normal density to ensure this is a true normal corrected from the Monte Carlo bias. We need to explicitly compute the weights. For each simulated point $X_i$, we compute it assumed density denoted by $d_i = \mathcal{N}(\hat{\mu}
,\hat{\Sigma})(X_i)$ where $\mathcal{N}(\hat{\mu},\hat{\Sigma})(.)$ denotes the p.d.f. of the multi-variate Gaussian. We divide these density by their sum to get weights $(w_i)_{i=1..k}$ that are positive and sum to one as follows. $w_j = d_j / \sum_{i=1}^k d_i$. Hence for $k$ simulated points, we get $\{X_i, w_i \}_{i = 1..k}$. We reorder jointly the uplets (points and density) in terms of their weights in decreasing order. To insist we take sorted value in decreasing order with respect to the weights $(w_i)_{i=1..k}$, we denote the order statistics $(i), w\downarrow$. This first sorting leads to k new  uplets $ \{X_{(i), w\downarrow}, w_{(i), w\downarrow}\}_{i = 1..k}$. Using a \emph{stable} sort (that keeps the order of the density), we sort jointly the uplets (points and weights)  according to their objective function value (in increasing order this time) and get a k new uplets $ \{X_{(i), f\uparrow}, w_{(i), w\downarrow}\}_{i = 1..k}$. We can now compute a new mean as follows:
\begin{equation}\label{eq:weighted_mean}
\hat{\mu} =  \underbrace{\sum_{i=1}^k  {w_{(i), w\downarrow}} \cdot  X_{(i), f\uparrow}}_{\mathrm{MC\,mean\,for\,} X_{f\uparrow}} - \underbrace{\left(\sum_{i=1}^k  w_i X_i - \hat{\mu} \right)}_{\mathrm{MC\,bias\,for\,} X}
\end{equation}
The intuition of equation \eqref{eq:weighted_mean} is to compute in the left term the Monte Carlo mean using reordered points according to their objective value and correct our initial computation by the Monte Carlo bias computed as the right term, equal to the initial Monte Carlo mean minus the real mean. We call this \textbf{strategy one}.

\item If we think for a minute about the strategy one, we get the intuition that when starting the minimization, it may not be optimal. This is because weights are proportional to $\exp( \frac{1}{2} (X-\hat{\mu})^T \hat{\Sigma}^{-1} (X-\hat{\mu}))$. When we start the algorithm, we use a large search space, hence a large covariance matrix $\hat{\Sigma}$ which leads to have weights which are quite similar. Hence even if we sort candidates by their fit, ranking them according to the value of $f$ in increasing order, we will move our theoretical multi variate Gaussian little by little. A better solution is more to brutally move the center of our multi variate Gaussian to the best candidate seen so far, as follows:
\begin{equation}\label{eq:simple_mean}
\hat{\mu} = \argmin_{X \in \mathcal{X}} f(X)
\end{equation} 
We call this \textbf{strategy two}. Intuitively, strategy two should be best when starting the algorithm while strategy one would be better once we are close to the solution.
\end{itemize}

To recover the true variance, we can adapt what we did in strategy one as follows:
\begin{itemize}
\item \begin{eqnarray}\label{eq:weighted_cov}
 \hat{\Sigma}  = & \underbrace{{\sum_{i=1}^k  {w_{(i), w\downarrow}} \cdot  \left(X_{(i), f\uparrow} - \overline{X}_{(.), f\uparrow}\right) \left(X_{(i), f\uparrow} - \overline{X}_{(.), f\uparrow}\right)^T }{}}_{\mathrm{MC\,covariance \,for\,} X_{f\uparrow}} \nonumber \\
&\!\!\!\!\!\!\!\! - \underbrace{\left(  {\sum_{i=1}^k  {w_{i}} \! \cdot \! \left(X_{i}- \overline{X} \right) \left(X_{i} - \overline{X} \right)^T  }{} \!- \!\hat{\Sigma}   \right)}_{\mathrm{MC\,covariance\,for\,simulated\,} X} \,\,\,\,\,
\end{eqnarray}
where $ \overline{X}_{(.), f\uparrow} = \sum_{i=1}^k w_{(i), w\downarrow}X_{(i), f\uparrow}$ and $\overline{X}=\sum_{i=1}^k w_{i} X_{i}$ are respectively the mean of the sorted and non sorted points.
\end{itemize}

\begin{algorithm}[!ht]
\caption{Predict and Correct parameters at step k} \label{algo:BayesianCMAES}
\begin{algorithmic} [1]
\STATE \textbf{Simulate candidate}
\STATE Use mean values $\hat{\mu}= \E[\mu]$ and $\hat{\Sigma}=\E[\Sigma]$
\STATE Simulate k points $\mathcal{X}= \{X_i \}=1..n \sim \mathcal{N}(\hat{\mu}, \hat{\Sigma})$
\STATE Compute densities $(d_i)_{i ..n} = (\mathcal{N}(\hat{\mu},\hat{\Sigma})(X_i) )_{i ..n} = $
\STATE Sort in decreasing order with respect to $d$ to get \\
$ \{X_{(i), d\downarrow}, d_{(i), d\downarrow}\}_{i = 1..n}$
\STATE Stable Sort in increasing order order with respect to $f(X_i)$ to get $\{X_{(i), f\uparrow}, d_{(i), d\downarrow}\}_{i = 1..n}$
\\ 
\STATE 
\STATE \textbf{Correct $\hat{\mu}$ and $ \hat{\Sigma}$}
\STATE Either Update $\hat{\mu}$ and $ \hat{\Sigma} $ using \eqref{eq:simple_mean} and \eqref{eq:weighted_cov}  \textbf{(strategy two)} 
\STATE Or Update $\hat{\mu}$ and $ \hat{\Sigma} $ using \eqref{eq:weighted_mean} and \eqref{eq:weighted_cov}  \textbf{(strategy one)} 
\STATE Update $\mu_{k+1},  \lambda_{k+1}, v_{k+1}, \psi_{k+1}$ using proposition \ref{prop:posterior_update}
\end{algorithmic}
\end{algorithm}

\section{Numerical results}\label{sec:experiments}
\subsection{Functions examined}
We have examined five functions to stress test our algorithm. They are listed in increasing order of complexity for our algorithm and correspond to different type of functions. They are all generalized function that can defined for any dimension $n$. For all, we present the corresponding equation for a variable $x=(x_1,x_2, .., x_n)$ of $n$ dimension. Code is provided in supplementary materials. We have frozen seeds to have \emph{reproducible of results}.

\subsubsection{Cone}
\begin{figure}[!ht]
\centering
\includegraphics[width=7cm]{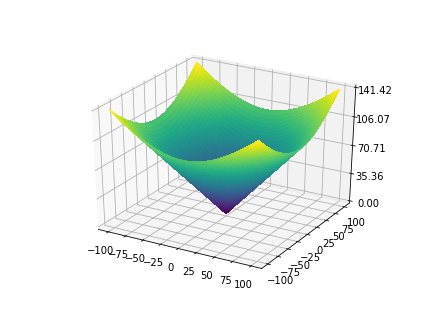}
\caption{A simple convex function: the quadratic norm. Minimum in 0} \label{fig:cone}
\end{figure}

The most simple function to optimize is the quadratic cone whose equation is given by \eqref{eq:function_Cone} and represented in figure \ref{fig:cone}. It is also the standard Euclidean norm. It is obviously convex and is a good test of the performance of an optimization method.
\begin{equation}\label{eq:function_Cone}
f(x)= \left( \sum_{i=1}^n x_i^2 \right)^{1/2} = \| x \|_2
\end{equation}

\subsubsection{Schwefel 2 function}
\begin{figure}[!ht]
\centering
\includegraphics[width=7cm]{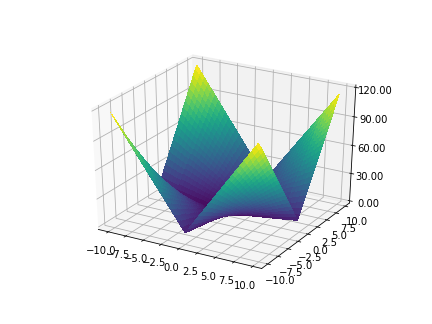}
\caption{Schwefel 2 function: a simple piecewise linear function} \label{fig:Schwefel2}
\end{figure}

A slightly more complicated function is the Schwefel 2 function whose equation is given by \eqref{eq:function_Schwefel2} and represented in figure \ref{fig:Schwefel2}. It is a piece wise linear function and validates the algorithm can cope with non convex function.

\begin{equation}\label{eq:function_Schwefel2}
f(x)= \sum_{i=1}^n \mid x_i\mid + \mathbb{P}od_{i=1}^n\mid x_i \mid
\end{equation} 

\subsubsection{Rastrigin}
\begin{figure}[!ht]
\centering
\includegraphics[width=7cm]{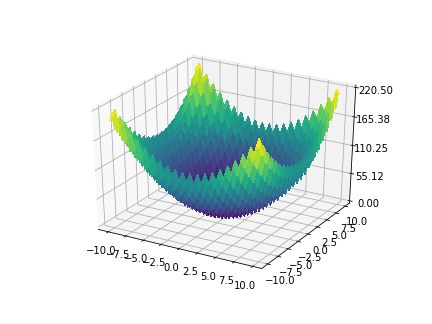}
\caption{Rastrigin function: a non convex function multi-modal and with a large number of local minima} \label{fig:Rastrigin}
\end{figure}

The Rastrigin function, first proposed by \cite{Rastrigin_1974} and generalized by \cite{Muhlenbein_1991}, is more difficult compared to the Cone and the Schwefel 2 function. Its equation is given by \eqref{eq:function_Rastrigin} and represented in figure \ref{fig:Rastrigin}. It is a non-convex function often used as a performance test problem for optimization algorithms. It is a typical example of non-linear multi modal function. Finding its minimum is considered a good stress test for an optimization algorithm, due to its large search space and its large number of local minima. 

\begin{equation}\label{eq:function_Rastrigin}
 f(x)=10 \times n + \sum_{i=1}^n \left[ x_i^2-10 \cos(2\pi x_i)\right]
\end{equation}

\subsubsection{Schwefel 1 function}
\begin{figure}[!ht]
\centering
\includegraphics[width=7cm]{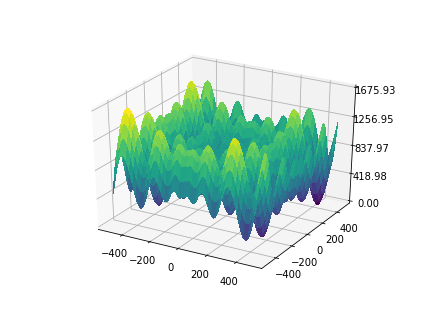}
\caption{Schwefel 1 function: a non convex function multi-modal and with a large number of local pronounced bowls} \label{fig:Schwefel1}
\end{figure}

The Schwefel 1 function whose equation is given by \eqref{eq:function_Schwefel1} is a tricky function to optimize. It is  represented in figure \ref{fig:Schwefel1}. 
It is sometimes only defined on $\left[-500, 500 \right]^n$. The Schwefel 1 function shares similarities with the Rastrigin function. It is continuous, not convex, multi-modal and with a large number of local minima. The extra difficulty compared to the Rastrigin function, the local minima are more pronounced local bowl making the optimization even harder.

\begin{eqnarray}\label{eq:function_Schwefel1}
&& \hspace{-0.5cm} f(x) = 418.9829\, \times \, n  \\
&& - \sum_{i=1}^n  \left[ x_i \sin(\sqrt{\mid x_i \mid}) \mathbbm{1}_{|x_i|< 500} + 500 \sin(\sqrt{500})  \mathbbm{1}_{|x_i| \geq 500} \right] \nonumber
\end{eqnarray}

\subsubsection{Eggholder function}
The Eggholder function whose equation is given by \eqref{eq:function_Eggholder} is a difficult function to optimize, because of the large number of local minima. 
It is sometimes only defined on $\left[-512, 512 \right]^n$. It shares similarities with the Schwefel1 function. It is continuous, not convex, multi-modal and with a large number of local minima. 
\begin{eqnarray}\label{eq:function_Eggholder}
f(x,y) = - \left(y+47\right) \sin \sqrt{\left|\frac{x}{2}+\left(y+47\right)\right|} - x \sin \sqrt{\left|x - \left(y + 47 \right)\right|}
\end{eqnarray}

\begin{figure}[!ht]
\centering
\includegraphics[width=7cm]{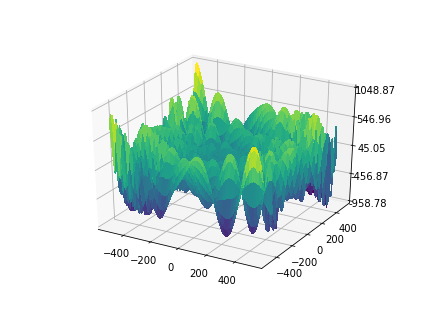}
\caption{Eggholder function: a non convex function multi-modal and with a large number of local pronounced bowls} \label{fig:Eggholder}
\end{figure}

\subsection{Convergence}
\begin{figure}[!ht]
\centering
\includegraphics[height=5.5cm]{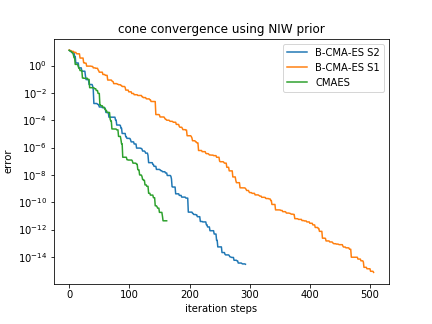}
\includegraphics[height=5.5cm]{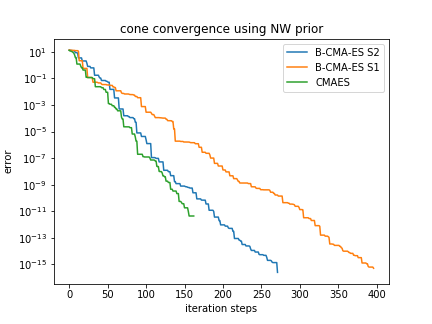}
\caption{Convergence for the Cone function. B-CMA-ES S2 outperforms standard CMA-ES and B-CMA-ES S1.}
\label{fig:convergence_cone}
\end{figure}

\begin{figure}[!ht]
\centering
\includegraphics[height=5.5cm]{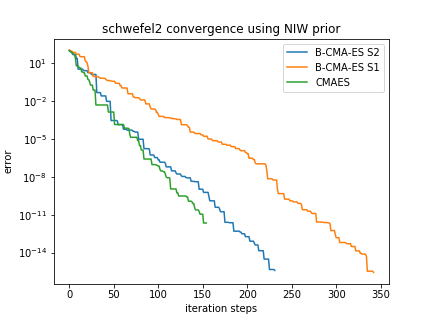}
\includegraphics[height=5.5cm]{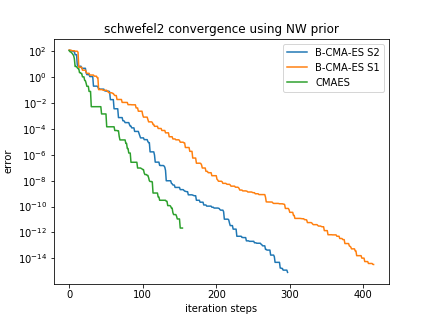}
\caption{Convergence for the Schwefel 2 function. B-CMA-ES S2 outperforms standard CMA-ES and B-CMA-ES S1.}
\label{fig:convergence_Schwefel2}
\end{figure}

\begin{figure}[!ht]
\centering
\includegraphics[height=5.5cm]{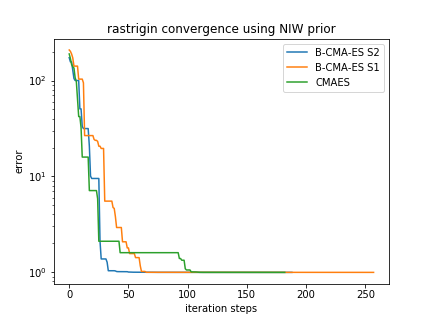}
\includegraphics[height=5.5cm]{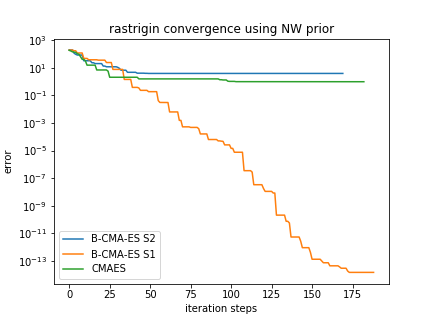}
\caption{Convergence for the Rastrigin function. B-CMA-ES S2 outperforms standard CMA-ES and B-CMA-ES S1.}
\label{fig:convergence_Rastrigin}
\end{figure}

\begin{figure}[!ht]
\centering
\includegraphics[height=5.5cm]{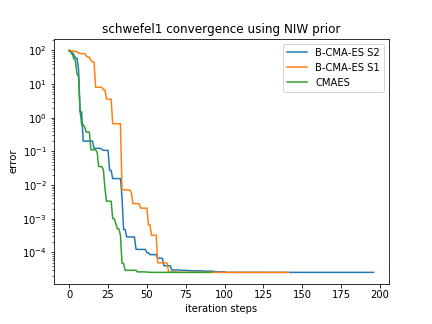}
\includegraphics[height=5.5cm]{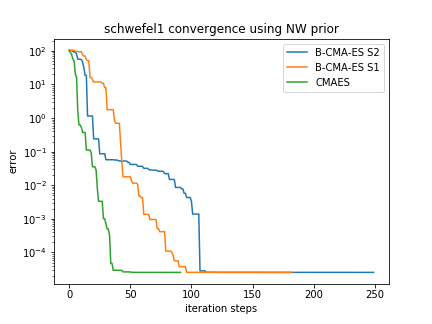}
\caption{Convergence for the Schwefel 1 function. B-CMA-ES S2 outperforms standard CMA-ES and B-CMA-ES S1.}
\label{fig:convergence_Schwefel1}
\end{figure}

For each of the functions, we compared our method using strategy one entitled \emph{B-CMA-ES S1}: update $\hat{\mu}$ and $ \hat{\Sigma} $ using \eqref{eq:weighted_mean} and \eqref{eq:weighted_cov} in \emph{orange} with strategy two \emph{B-CMA-ES S2}: same update but using \eqref{eq:simple_mean} and \eqref{eq:weighted_cov}, in \emph{blue} and standard CMA-ES as provided by the opensource python package pycma in \emph{green}. We clearly see that strategy two outperforms standard CMA-ES and Bayesian CMA-ES S1. The convergence graphics that shows the error compared to the minimum are represented 
\begin{itemize}
\item for the cone function by figure \ref{fig:convergence_cone} (case of a convex function), with initial point $(10,10)$
\item for the Schwefel 2 function in figure \ref{fig:convergence_Schwefel2} (case of piecewise linear function), with initial point $(10,10)$
\item for the Rastrigin function in figure \ref{fig:convergence_Rastrigin} (case of a non convex function with multiple local minima), with initial point $(10,10)$
\item and for the Schwefel 1 function in figure \ref{fig:convergence_Schwefel1} (case of a non convex function with multiple large bowl local minima), with initial point $(10,10)$
\end{itemize}

\begin{figure}[!ht]
\centering
\includegraphics[height=5.5cm]{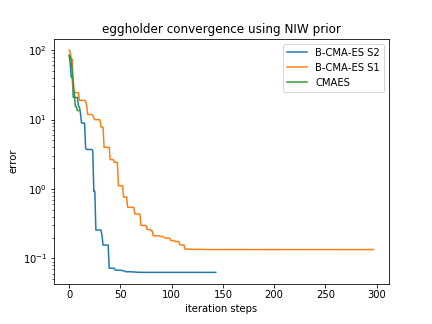}
\includegraphics[height=5.5cm]{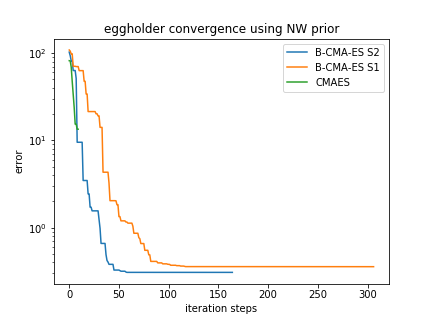}
\caption{Convergence for the Eggholder function}
\label{fig:convergence_Eggholder}
\end{figure}

For functions that are convex, our method performs similarly as standard CMA-ES. For function with harder local minima, the Bayesian CMA-ES is able to perform better. We conjecture that this is due to contraction dilatation mechanism that enables to avoid being trapped in a local minimum.

\section{Conclusion}
In this paper, we have revisited the CMA-ES algorithm and provided a Bayesian version of it. Taking conjugate priors, we can find optimal update for the mean and covariance of the multi variate Normal. We have provided the corresponding algorithm that is a new version of CMA-ES. First numerical experiments show this new version is comparable to standard CMA-ES on traditional functions such as cone, Schwefel 1, Rastrigin and Schwefel 2. The similar convergence can be explained on a theoretical side from the optimal update of the prior (thanks to Bayesian update) and the use of the best candidate seen at each simulation to shift the mean of the multi-variate Gaussian likelihood. We envisage further works to benchmark our algorithm to traditional CMA-ES and other evolutionary algorithms, in particular to use the COCO platform to provide more meaningful tests and confirm the theoretical intuition of good performance of this new version of CMA-ES, and to test the importance of the prior choice. 

\section{Appendix}

\subsection{Conjugate priors}\label{proof:conjugate_priors}
\begin{proof}
Consider $n$ independent and identically distributed (IID) measurements $\mathcal{X}\triangleq\{ \x^j\in \mathbb{R}^d | 1\leq j \leq n\}$ and assume that these variables have an exponential family density. The likelihood $p(\mathcal{X}|\theta,\kappa)$, writes simply as the product of each individual likelihood:
\begin{align}
\hspace{-0.3cm} p(\mathcal{X}|\theta,\kappa) \! = \!\Big(\prod_{j=1}^{n}h(\x^j)\Big) \exp\Big(\eta(\theta,\kappa)^T \sum_{j=1}^n T(x^j)-n A(\eta(\theta,\kappa))\Big).  \label{eq:ExponentialFamilyLikelihood}
\end{align}
If we start with a prior $\pi(\theta)$ of the form $\pi( \theta )\propto \exp (\mathcal{F}(\theta))$ for some function $\mathcal{F}(\cdot)$, its posterior writes: 
\begin{align}
\pi( \theta | \mathcal{X} ) &\propto p( \mathcal{X} | \theta )  \exp (\mathcal{F}(\theta)) \nonumber \\
&\propto  \exp \left( \eta(\theta, \kappa) \cdot \sum_{j=1}^n T(x^j)-n A(\eta(\theta,\kappa)) +\mathcal{F}(\theta)\right).\label{eq:ExponentialFamilyPosterior-expanded}
\end{align}
It is easy to check that the posterior \eqref{eq:ExponentialFamilyPosterior-expanded} is in the same exponential family as the prior iff $\mathcal{F}(\cdot)$ is in the form
\begin{align}
\mathcal{F}(\theta)= \lambda_1\cdot \eta(\theta,\kappa) - \lambda_0 A(\eta(\theta,\kappa))
\end{align} 
for some $\lambda\triangleq(\lambda_0, \lambda_1)$, such that 
\begin{align}
\!\!\!\! p(\mathcal{X}|\theta,\kappa)  \! \propto \! \exp\Big(\Big(\lambda_1+ \sum_{j=1}^n T(x^j)\Big)^T \!\!  \eta(\theta,\kappa)  -( n +\lambda_0) A(\eta(\theta,\kappa)) \Big)\! .\label{eq:ExponentialFamilyPosteriorLinearcombination}
\end{align}
Hence, the conjugate prior for the likelihood \eqref{eq:ExponentialFamilyLikelihood} is parametrized by $\lambda$ and  given by 
\begin{align}
p(\mathcal{X}|\theta,\kappa)  =\frac{1}{Z}\exp\left( \lambda_1\cdot \eta(\theta,\kappa) - \lambda_0 A(\eta(\theta,\kappa)) \right),
\end{align} 
where $Z={\int{\exp\left(\lambda_1\cdot \eta(\theta,\kappa) - \lambda_0 A(\eta(\theta,\kappa)) \right) \d x}}$.
\end{proof}

\subsection{Multivariate Canonical form}\label{proof:one_line}
In the case of the multi variate normal, the canonical form for this distribution writes as
\begin{eqnarray}\label{eq:normal_multivariate}
\hspace{-1cm} & & \frac{1}{\sqrt{(2 \pi)^d \det(\Sigma)}} \exp \left(-  \frac { (X-\mu)^T \Sigma^{-1} (X-\mu)}{2} \right) \nonumber \\
\hspace{-1cm} &= & \exp\left( ( \Sigma^{-1} \mu,  \Sigma^{-1} )^T \cdot ( X, - \frac 1 2  \vect( X X^{T}) \right)  \nonumber\\
\hspace{-1cm} & & \hspace{1cm} \frac{1}{(2 \pi)^{d/2} } \exp \left( -\frac 1 2 \mu^T \Sigma^{-1} \mu - \frac 1 2 \log\left(\det(\Sigma)\right)  \right) 
\end{eqnarray}

which gives the following moment and canonical parameters:
\begin{eqnarray}\label{eq:normal_multivariate2}
\theta 			& =& (\mu, \Sigma) \nonumber \\
T(X) 			&= &\left( X, -\frac 1 2 \vect( X X^{T} ) \right)  \nonumber  \\
\eta(\theta ) 		&=& \left(\Sigma^{-1} \mu,  \Sigma^{-1} \right)  \nonumber  \\
A(\eta(\theta ))	&=& \frac 1 2 \mu^T \Sigma^{-1} \mu + \frac 1 2 \log( \det(\Sigma)) \nonumber  \\
h(x)				&=& \frac{1}{(2 \pi)^{d/2} } \label{eq:normal_exponential_family}
\end{eqnarray}

\subsection{Conjugate priors determination}\label{proof:conjugate_prior_normal}
Using proposition \ref{prop:conjugate_priors} and the exponential family formulation of the multi variate normal (equations \eqref{eq:normal_exponential_family}), we have that any conjugate prior for the multi variate normal that belongs to the exponential family is given by
\begin{eqnarray}
\pi(\theta)  &\propto &\exp\left( \lambda_1\cdot \eta(\theta) - \lambda_0 A(\eta(\theta)) \right) \nonumber  \\
&\propto & \exp\left( \lambda_1\cdot \left(\Sigma^{-1} \mu,  \Sigma^{-1} \right)   \right) \nonumber  \\ 
& & \hspace{0.2cm}  \exp\left( -\frac 1 2 \mu^T (\frac{\Sigma}{\lambda_0})^{-1} \mu - \frac 1 2 \log( \det(\frac{\Sigma}{\lambda_0})) \right) 
\end{eqnarray}
If we write $ \lambda_1 = ( \lambda_0 \, \mu_0,  \lambda_2) $ and $\Psi_0 = -2  (\lambda_2+ \frac{\lambda_0}{2} \mu_0 \mu_0^T )$, we get
\begin{eqnarray}
\pi(\theta)  &\propto & \exp\left( -\frac 1 2 (\mu-\mu_0) ^T (\frac{\Sigma}{\lambda_0})^{-1} (\mu-\mu_0) - \frac 1 2 \log( \det(\frac{\Sigma}{\lambda_0})) \right)   \nonumber  \\ 
& & \hspace{0.2cm}  \exp\left( - \frac 1 2  \Psi_0 \cdot \Sigma^{-1} \right) 
\end{eqnarray}

The first term is a normal multi variate distribution. 
Its  parameters are $\mu_0$ and $\frac{\Sigma}{\lambda_0}$. 

In the second term, we can recognize the proportional term of an inverse Wishart $\exp\left( - \frac 1 2 \Tr( \Psi_0 \Sigma^{-1} ) \right)$, with parameters $\nu_0, \Psi$. 

This shows the conjugate prior of the multi variate normal given by its mean vector $\mu$ and covariance matrix $\Sigma$ is a normal inverse Wishart. 
Its parameters are $NIW( \mu_0, \lambda_0,\nu_0, \Psi_0)$  \qed

If the multi variate normal is parametrized by its mean vector $\mu$ and its precision matrix $\Lambda$, the same reasoning gives
\begin{eqnarray}
\pi(\theta)  &\propto &\exp\left( \lambda_1\cdot \left(\Lambda \mu, \Lambda \right)   \right) \nonumber  \\ 
& & \hspace{0.2cm}  \exp\left( -\frac 1 2 \mu^T (\lambda_0 \Lambda) \mu + \frac 1 2 \log( \det(\frac{\Lambda}{\lambda_0})) \right) 
\end{eqnarray}

The second term is a multi variate normal distribution given by 
$N( \mu_0, (\lambda_0 \Lambda)^{-1} )$ while the first one is the term of a Wishart distribution 
that is proportional to $exp( \frac 1 2 \Tr( W^{-1} \Lambda )$ whose parameters are $\mathcal{W}( W_0,  \nu_0)$. 
This shows that the conjugate prior of the multi variate normal described by its mean vector $\mu$ and 
precision matrix $\Lambda$ is a normal Wishart distribution $NW( \mu_0, \lambda_0, \nu_0, W_0)$ \qed

\subsection{Posterior update}\label{proof:conjugate_prior_computation}
The posterior update is quite straightforward and very similar for the two cases: NIW and NW. We will detail only the calculation for the NIW case as it is very similar for the NW. 
Recall that the probability density function of a Normal inverse Wishart random variable is expressed as the product of a Normal and an Inverse Wishart probability density functions. 
Denoting by $p \times p $ the dimension of the covariance matrix $\Sigma$ and using the Bayes rules, the posterior is proportional to the product of the prior and likelihood:
\begin{eqnarray}
& & \mathrm{posterior} \nonumber \\
&\propto  &\mathrm{prior} \times \mathrm{likelihood} \nonumber \\
&\propto & \sqrt{\frac{\lambda_0}{|\Sigma|}} \exp(- \frac{1}{2} (\mu-\mu_0)^T (\frac{\Sigma}{\lambda_0})^{-1}  (\mu-\mu_0) )   \nonumber \\
& & \hspace{0.2cm} \times |\Psi|^{\nu_0/2} |\Sigma|^{-\frac{\nu_0+p+1}{2}} \exp( -\frac{1}{2}\Tr( \Sigma^{-1}  \Psi_0 )) \nonumber \\
& & \hspace{0.2cm} \times \,  |\Sigma|^{-n/2}   \prod_{i=1}^{n} \exp(- \frac 1 2 (x_i-\mu )^T \Sigma^{-1}  (x_i-\mu) ) \hspace{1.5cm}
\end{eqnarray}
First of all, we can regroup all terms in $x_i$ as follows
\begin{eqnarray}
 \hspace{-2.3cm} & \prod_{i=1}^{n} \exp(- \frac 1 2 (x_i-\mu )^T \Sigma^{-1}  (x_i-\mu) )  \nonumber \\
 \hspace{-2.3cm} = & \!\!\!\!  \exp( - \frac 1 2  \sum_{i=1}^{n} (x_i-\mu )^T \Sigma^{-1}  (x_i-\mu) ) 
\end{eqnarray}
\noindent and use the following remarkable identity:
\begin{eqnarray}
\hspace{-0.75cm} & &   \sum_{i=1}^n \left( x_i-\mu\right)^T \Sigma^{-1} \left( x_i-\mu\right) \nonumber \\
\hspace{-0.75cm} & = & n \bigg[ \frac{1}{n} \sum_{i=1}^n \left( x_i-\overline{x}\right)^T \Sigma^{-1} \left( x_i-\overline{x}\right) +  \left( \overline{x}-\mu \right)^T \Sigma^{-1} \left( \overline{x}-\mu \right) \bigg]  \hspace{0.75cm}  \nonumber \\
\hspace{-0.5cm} & = & \Tr( \Sigma^{-1} \, n C) + n  \left( \overline{x}-\mu \right)^T \Sigma^{-1} \left( \overline{x}-\mu \right)
\end{eqnarray}
\noindent where we have used the commutativity property of the trace operator $\Tr(A B ) = \Tr(B A )$ and that for a real number, the number is equal to its trace and written $C =  \frac{1}{n} \sum_{i=1}^n \left( x_i-\overline{x}\right)^T \Sigma^{-1} \left( x_i-\overline{x}\right)$ the sample covariance. Going further, we have

\begin{eqnarray}
\hspace{-0.5cm} & &   (\mu-\mu_0)^T  (\frac{\Sigma}{\lambda_0})^{-1}  (\mu-\mu_0) + n  \left( \overline{x}-\mu \right)^T \Sigma^{-1} \left( \overline{x}-\mu \right) \nonumber \\
\hspace{-0.5cm} & = \!\! & (\lambda_0+n)   \mu ^T \Sigma^{-1} \mu -2 (\lambda_0 \, \mu_0 + n  \overline{x} )^T \Sigma^{-1} \mu + \lambda_0 
\mu_0^T  \Sigma^{-1}  \mu_0  \nonumber  \\
& & + n  \overline{x}^T \Sigma^{-1} \overline{x} 
\end{eqnarray}

\begin{eqnarray}
\hspace{-0.5cm} & = \!\! & (\lambda_0+n)  \left( \mu - \frac{\lambda_0 \mu_0+n  \overline{x}}{\lambda_0+n} \right)^T \Sigma^{-1} \left( \mu - \frac{\lambda_0 \mu_0+n  \overline{x}}{\lambda_0+n} \right) \nonumber \\
\hspace{-0.5cm} & &   \hspace{0.2cm} - \frac{1}{\lambda_0+n}  \left(\lambda_0 \mu_0+n  \overline{x}\right)^T \Sigma^{-1} \left(\lambda_0 \mu_0+n  \overline{x} \right) \nonumber \\
\hspace{-0.5cm} & &   \hspace{0.2cm} + \,\, \lambda_0 \, \mu_0^T   \, \Sigma^{-1}   \, \mu_0 + n   \, \overline{x}^T  \, \Sigma^{-1}  \, \overline{x}
\end{eqnarray}

\begin{eqnarray}
&=& (\lambda_0+n)  \left( \mu - \frac{\lambda_0 \mu_0+n  \overline{x}}{\lambda_0+n} \right)^T \!\!\Sigma^{-1} \left( \mu - \frac{\lambda_0 \mu_0+n  \overline{x}}{\lambda_0+n} \right)\nonumber \\
& & + \Tr( \Sigma^{-1} n D )
\end{eqnarray}
\begin{eqnarray}
&  \mathrm{where } \hspace{1 cm}  D= \frac{\lambda_0 \, n}{n(\lambda_0+n)} ( \overline{x}-\mu_0) ( \overline{x}-\mu_0)^T
\end{eqnarray}

Hence, we can compute explicitly the posterior as follows:
\begin{eqnarray}
\hspace{-1cm}& &\mathrm{posterior} \nonumber \\
\hspace{-1cm}&\propto &\sqrt{\frac{\lambda_1}{|\Sigma|}} \exp\left\{- \frac{1}{2} \big(\mu-\mu_1 \big)^T
  (\frac{\Sigma}{\lambda_1})^{-1}  \big(\mu-\mu_1 \big) \right\}\nonumber \\
\hspace{-1cm} & &\hspace{0.5cm} \times|\Psi_1|^{\nu_1 / 2} |\Sigma|^{-\frac{\nu _1+p+1}{2}} \exp\left\{- \frac{1}{2} \Tr\left( \Sigma^{-1} \Psi_1 \right) \right\}
\end{eqnarray}

\begin{eqnarray}
\mathrm{with } \quad \mu_1 	  & = & \frac{\lambda_0 \, \mu_0 + n \overline{x}}{\lambda_0+n}  \nonumber \\
\lambda_1 & = & \lambda_0 + n  \nonumber \\
\nu_1	  & = & \nu_0 + n \nonumber \\
\Psi_1 	  & = & \Psi_0 + n C +n D \hspace{2cm}
\end{eqnarray}

which are exactly the equations provided in \eqref{eq:update_posterior_niw} \qed

\subsection{Convexity of the inverse of a matrix}\label{proof:inverse_matrix}
We give here \textbf{six different proofs} of the convexity of the inverse of a matrix in the domain of symmetric definite positive matrices $\mathcal{S}^p_{++}$. The first and second proofs relies on the fact that the result is a consequence of proving that the matrix fractional function $f(X,y) = y^T X^{-1 }y$ is convex on the domain $\mathrm{dom} f = \mathcal{S}^p_{++} \times \mathbb{R}^p$. The implication comes from the fact that 
\begin{align}\label{eq:proof1_1}
		& f \text{ is convex } \nonumber \\
\implies 	& (1-\lambda) f( M,y) + \lambda f(N,v) \geq f( (1-\lambda) M + \lambda N, y ) \nonumber \\
\implies	& y^T \left[ (1-\lambda) M^{-1} + \lambda N^{-1} - ((1-\lambda) M + \lambda N)^{-1}\right] y \ge 0
\end{align}
Since $y$ is arbitrary, this implies the matrix within the square bracket in equation \eqref{eq:proof1_1} is positive semi-definite. It is interesting to notice that matrix fractional function is in a sense an extension of the fact that the quadratic over linear function defined as $f(x,y)=x^2/y$ is convex on $\mathbb{R}_+^2$.

\begin{proof}
The \textbf{first proof} uses the property that the minimum of a convex function over a convex set is convex. For $\Sigma \in S_{++}^n$, and for $u, y \in \mathbb{R}^n$ we can consider the quadratic function $f(u)$ defined by
\[
f(u) = \frac{1}{2} u^{T} \Sigma u - u^T y
\]
As  $\Sigma \in S_{++}^n$, this function is a obviously convex (quadratic function with its quadratic coefficient given by a definite positive matrix). Hence its minimum  $ \inf_{u \in \mathbb{R}^n} f(u) $ over a convex set  is convex.
 Its easy to minimize a quadratic function and find its minimum given by the stationary point of its gradient $\frac{1}{2}  y^{T} \Sigma^{-1} y$, which concludes the proof.
\end{proof}

\begin{proof}
A \textbf{second proof} is to show that the epigraph of $f$, denoted by $\text{epi}(f)$ is convex thanks to the link between positive semi definite cones and Schur complements. We have that 
\begin{eqnarray}
x \in \text{epi}(f)  \Leftrightarrow  y^{T} \Sigma^{-1} y \leq t \Leftrightarrow \left[  \begin{array}{cc}  \Sigma & y \\ y^T & t \end{array} \right]  \succeq 0 
\end{eqnarray}
This concludes the proof as the epigraph of $f$ is convex as the inverse image of the positive semi definite cone $S_{++}^{n+1}$ by the Schur complement that is an affine mapping.  
\end{proof}

\begin{proof}
A \textbf{third proof} relies on the fundamental identity of the inverse of a matrix $X$: $X X^{-1} = I_p$, where $I_p$ is the identity matrix with $p$ rows (or columns).
Take $M, N$ two positive definite symmetric matrices and $\lambda \in [0,1]$.
Take $P_{\lambda} = (1-\lambda) M + \lambda N$. $P$ and $P^{-1}$ are obviously symmetric positive definite. 
Denote by $(.)'$ the derivative with respect to $\lambda$. We have:

$$
P P^{-1} = I_p
\implies P'P^{-1} + P ( P^{-1} )' = 0_p
\implies (P^{-1})' = - P^{-1} P' P^{-1}
$$
Notice that $P'' = 0_p$, since $P$ is linear in $\lambda$. Differentiate one more time to get:
\begin{equation}\label{eq:proof3_1}
(P^{-1})'' = - (P^{-1})' P' P^{-1} - P^{-1}P' (P^{-1})' = 2 P^{-1}P' P^{-1} P' P^{-1} 
\end{equation}

For any non-zero random vector $y$, define $v_{\lambda} = P'_{\lambda} P^{-1}_{\lambda} y$ and $\varphi_{\lambda} = y^T P^{-1}_{\lambda}  y$. Equations \eqref{eq:proof3_1} says that 
\begin{equation}\label{eq:proof3_2}
\varphi''_{\lambda} = y^T (P^{-1}_{\lambda})'' y = 2 {v^T}_{\lambda} P^{-1}_{\lambda} v_{\lambda} \ge 0
\end{equation}

since $P^{-1}_{\lambda}$ is positive definite. As the second order derivative is positive, we conclude that $\varphi_{\lambda}$ is a convex function for $\lambda$ over $[0,1]$. 
As a result, for any $\lambda \in (0,1)$, we have:

\begin{align}\label{eq:proof3_3}
	&(1-\lambda)\varphi(0) + \lambda\varphi(1) - \varphi_{\lambda} \ge 0 \nonumber \\
\iff	& y^T \left[ (1-\lambda) M^{-1} + \lambda N^{-1} - ((1-\lambda) M + \lambda N)^{-1}\right] y \ge 0
\end{align}

Since $y$ is arbitrary, this implies the matrix within the square bracket in \eqref{eq:proof3_3} is positive semi-definite and hence:
$$(1-\lambda) M^{-1} + \lambda N^{-1} \succeq ((1-\lambda) M + \lambda N)^{-1}$$

Please note that when $P' = N - M$ is invertible, $v_{\lambda}$ is non-zero for non-zero $y$. The inequalities in \eqref{eq:proof3_2} and \eqref{eq:proof3_3}  become strict and the matrix within the square bracket in \eqref{eq:proof3_3}  is positive definite instead of positive semi-definite.
\end{proof}

\begin{proof}
A \textbf{fourth proof} is to derive the convexity of the inverse of a matrix from the convexity of the function $f(t)=\frac 1 t$ for $t \geq 0$. Let $P= {X}^{-1/2} {Y} {X}^{-1/2}$. We want to prove that
\begin{eqnarray}
\hspace{-0.5cm}  && (1-\lambda)X^{-1}+\lambda Y^{-1} - ((1-\lambda)X+\lambda Y)^{-1}\succeq0 \label{eq:proof4_eq1} \\
\hspace{-0.5cm}  &\iff& \hspace{-0.3cm} X^{1/2}\left[(1-\lambda)X^{-1} \!+ \! \lambda Y^{-1} \!- \! ((1-\lambda)X \! + \! \lambda Y)^{-1}\right]X^{1/2}\succeq 0 \quad  \\
\hspace{-0.5cm}  &\iff& \hspace{-0.3cm} (1-\lambda)I+\lambda P^{-1}-((1-\lambda)I+\lambda P)^{-1}\succeq 0 
\end{eqnarray}
where in inequality \eqref{eq:proof4_eq1}, we have left- and right- multiplied both sides by ${X}^{1/2}$. As $ P$ is positive definite, it can be unitary diagonalised and hence without loss of generality, we can assume that it is a diagonal matrix. So, the inequality reduces down to the scalar case $(1-\lambda)+\lambda p_{ii}^{-1} \ge ((1-\lambda)+\lambda p_{ii})^{-1}$, which is true using the fact that the function $f(t)=\frac 1 t$ is convex for $t \geq 0$
\end{proof}

The last two proofs relies on the fact that the result is also implied by the fact that the function $f(X) = \Tr(X^{-1}yy^{t}) = \Tr(y^{t} X^{-1}y)$ is convex for $X \in \mathcal{S}^p_{++}$ for any $y \in  \mathbb{R}^p$. This comes from the nice property that the Trace operator can commute and that the trace of a real number is itself.

\begin{proof}
The \textbf{fifth proof} uses the fact that a positive second order derivative along any line is enough to prove convexity. Consider $S(t) = U + t V$ where $U$ and $V$ are symmetric positive definite.  It is enough to show that $ \left.\dfrac{d^2}{d t^2} \text{Tr}(y^{t} S(t)^{-1} y )\right|_{t=0} \ge 0$  
 We have 
\begin{eqnarray}
 S(t)^{-1} &=& (U (I + t U^{-1} V))^{-1} \nonumber \\
 &=& U^{-1} - t U^{-1} V U^{-1} + t^2 U^{-1} V U^{-1} V U^{-1} + \ldots
\end{eqnarray}
 So 
 \[   \left. \dfrac{d^2}{\partial t^2} \text{Tr}(S(t)^{-1}) \right|_{t=0} = 2 \text{Tr}( U^{-1} V U^{-1} V U^{-1})
 \]
 But $U^{-1} V U^{-1} V U^{-1} = W U^{-1} W^T$ where $W= U^{-1} V$ and $U^{-1}$ is positive definite, so $W U^{-1} W^T$ is positive semi definite, which implies 
$\text{Tr}(W U^{-1} W^T) \ge 0$, which concludes the proof
\end{proof}

\begin{proof}
A final \textbf{sixth proof}  is to relate this to eigen values. We can notive that the function $f(X) = \Tr(X^{-1}yy^{t})$ is indeed the sum of the inverse of eigen values denoted by $\lambda_i$.
\[
Tr((y^T X^{-1} y) = \sum \limits_{i=1..n} \frac{1}{ \lambda_i}
\]
We know that the function that associates to a diagonal matrix with strictly positive terms its kth element (which turns out to be one of its eigen values but not necessarily its kth one) is linear, hence convex and concave. By the composition rules for convex function, with $g(x)=1/x$, we can conclude that the inverse of the kth elements is convex for diagonal matrices with strictly positive term. Thus, the sum of the inverse of eigen values (defined as a sum of convex functions) is convex on the set of diagonal matrix with strictly positive term. We can conclude using the diagonalisation result of definite positive matrix (with $S=U D U^T$ , $U$ an orthonormal matrix, $D$ a diagonal matrix with strictly positive term and $S \in S_{++}^{n}$) to extend the convexity property to the set of $ S_{++}^{n}$ and use also that $Tr(AB)=Tra(BA)$  
\end{proof}

\clearpage
\normalsize
\bibliographystyle{apalike}
\bibliography{bibfile} 

\end{document}